\documentclass[pmlr]{jmlr}% new name PMLR (Proceedings of Machine Learning)

\RequirePackage{graphicx}
\usepackage{booktabs}
\usepackage{longtable}% for long tables
\makeatletter
\def\set@curr@file#1{\def\@curr@file{#1}} %temp workaround for 2019 latex release
\makeatother
\usepackage[load-configurations=version-1]{siunitx} % newer version
\usepackage{booktabs}
\usepackage{graphicx}
\newcommand{\hicdgraph}{\texttt{HICD-Graph}} %kat: would removing the whtiespace help with the weird whitespace after HICD-Graph in particular when followed by a comma?
\newcommand{\hicdbert}{\texttt{HICD-BERT}}
\theorembodyfont{\upshape}
\theoremheaderfont{\scshape}
\theorempostheader{:}
\theoremsep{\newline}

\usepackage{multirow}
\usepackage[table]{xcolor}

\title[Hierarchical Modeling of ICD Codes]{Hierarchical Modeling of ICD Codes in EHR
Foundation Models}

\author{\Name{Megha Thukral\nametag{$^{1}$}} \Email{mthukral3@gatech.edu} \AND
\Name{Dong Gyun Kang\nametag{$^{1}$}} \Email{dkang335@gatech.edu}\AND
\Name{Rudra Pratap Singh\nametag{$^{1}$}} \Email{rudra.singh@gatech.edu}\AND
\Name{Shruthi Kashinath Hiremath\nametag{$^{2}$}} \Email{shruthi.hiremath@optum.com}\AND
\Name{Katrin H\"ansel\nametag{$^{2}$}} \Email{katrin\_haensel@optum.com}\AND
\Name{Thomas Pl\"otz\nametag{$^{1}$}} \Email{thomas.ploetz@gatech.edu}\AND
\addr
$^{1}$ School of Interactive Computing, Georgia Institute of Technology,
Atlanta, United States\\
$^{2}$ Optum AI, United States
}

\begin{document}

\maketitle

\begin{abstract}
Electronic health record foundation models typically treat ICD diagnosis codes as flat tokens, overlooking the clinically meaningful hierarchical structure that captures disease families, subcategories, and fine-grained diagnostic detail. As a result, existing EHR representation learning methods do not explicitly exploit the hierarchical structure already present in the coding system. In this work, we study ICD-10-CM hierarchy as a \emph{general inductive bias} for clinical representation learning. 
We investigate two complementary mechanisms for incorporating hierarchy: first, by augmenting diagnosis sequences in a BERT-style transformer with tokens corresponding to different levels of the ICD hierarchy, and second, by injecting hierarchy into graph-based code representations through hierarchy-aware edges combined with diagnosis co-occurrence structure. Across these settings, we evaluate whether explicit hierarchy improves downstream prediction, which levels of the hierarchy are most useful, whether hierarchy encoding improves transfer across datasets, and how hierarchy reshapes embedding similarity structure. We conduct experiments on two large-scale real-world clinical datasets: MIMIC-IV, used for pretraining and in-domain evaluation, and eICU, used to assess cross-dataset transfer via frozen encoder probing. Our findings show that explicitly encoding ICD hierarchy improves over flat code representations in both in-domain and cross-dataset settings, while revealing that the most useful level of hierarchy depends on both the task and the modeling approach. More broadly, we focus on hierarchy-aware EHR representation learning and show that the benefits of encoding hierarchy are generalizable across modeling settings and hierarchy levels. Our code is available \href{https://meghathukral.github.io/HICD_EHR_FMs/}{here}.

\end{abstract}

\section{Introduction}
Electronic health records (EHRs) provide a longitudinal view of patients' clinical history through diagnoses, procedures, medications, and encounters.
Advances in transformer-based EHR models, such as BEHRT and Med-BERT, have shown that large-scale pretraining can learn useful patient representations for downstream clinical prediction tasks~\citep{li2020behrt,rasmy2021medbert}. 
However, much of this progress still treats structured clinical codes as flat symbols, even when those codes are drawn from ontologies that were explicitly designed to encode clinical hierarchy. 
This creates a mismatch between the structure of the data and the structure presented to the model.  
For instance, ICD codes are not arbitrary identifiers: they organize diseases into clinically meaningful families, subgroups, and increasingly specific categories~\citep{who_icd10} (see Figure~\ref{fig:icd_code}).
A code, therefore, carries both a fine-grained diagnosis and a path through a broader clinical taxonomy. 
In practice, however, most EHR models represent ICD codes as independent tokens, ignoring this hierarchical structure. 
As a result, related diagnoses that should share information may instead be learned as isolated symbols. 
This is especially problematic in clinical data, where many diagnoses are sparse, long-tailed, and institution-specific~\citep{choi2016doctor}.

We hypothesize that explicitly modelling hierarchy could benefit EHR representation learning in several ways. 
First, the hierarchy provides a natural inductive bias for relationships between diseases, so rare diagnosis codes can benefit from data associated with clinically similar codes~\citep{perotte2014diagnosis}.
Second, broader and intermediate groupings may capture care pathways or anatomical systems that are more predictive than the most specific code alone~\citep{song2019medical}. 
Third, hierarchy can improve robustness under \emph{domain shift}, since higher-level disease families are often more stable across institutions~\citep{ostrominski2021coding}. 
More broadly, encoding hierarchy can help learned embeddings better reflect clinical semantics rather than only empirical co-occurrence.

Prior work has explored incorporating ICD code hierarchy into healthcare representation learning, such as  GRAM ~\citep{choi2017gram}, primarily through graph-based approaches that embed ontological structure into code representations. However, these methods predate the current generation of transformer-based EHR foundation models, and it remains underexplored how the ICD hierarchy should be systematically incorporated into such models, and which granularity of hierarchy proves most beneficial.

To address this gap, we propose and evaluate two complementary strategies for injecting ICD hierarchy into contemporary EHR representation learning.
The first, \hicdbert \ (\textbf{H}ier\-archical \textbf{ICD}-BERT), injects hierarchy directly into token representations in a BERT-style encoder, embedding each level of the ICD hierarchy as an additive token embedding alongside the diagnosis code.
The second, \hicdgraph \ (\textbf{H}ierarchical \textbf{ICD}-Graph), encodes hierarchy relationally via a diagnosis co-occurrence graph augmented with ontology-derived edges, learning hierarchy-aware code embeddings that initialise a patient-level Transformer.
Together, these approaches allow us to compare whether hierarchical information is best consumed as a token-level signal or as a relational structure over the code vocabulary.

Our results show that hierarchy is a \textit{robustly effective inductive bias} for EHR representation learning, as across both architectures and both prediction tasks, the large majority of hierarchy-augmented configurations outperform their no-hierarchy baselines. 
Both models benefit from hierarchy, with graph-based encoding leveraging all hierarchy levels and token-based encoding benefiting most from the finest granularity level. 
Crucially, hierarchy also improves cross-dataset transfer, specifically for graph-based hierarchy encoding, which transfers robustly from MIMIC-IV to eICU.

\begin{itemize}
    \item We investigate ICD code hierarchy as an inductive bias for EHR representation learning across two architectures, two clinical prediction tasks, and a systematic ablation over three granularity levels, showing that hierarchy significantly improves performance in 26 of 28 comparisons. 
    
    \item We examine the benefit of hierarchy under distributional shift by evaluating cross-dataset transfer from MIMIC-IV to eICU under a frozen-encoder probe protocol, showing that graph-based hierarchy encoding transfers robustly to new datasets and task settings.

    \item Through embedding analysis of the learned code representations, we show that hierarchy produces more coherent code clusters, and that the configurations with the tightest clusters also achieve the strongest downstream performance.
\end{itemize}

\subsection*{Generalizable Insights about Machine Learning in the Context of Healthcare} 
Our findings suggest two broader lessons for machine learning in health. First, structured clinical ontologies such as ICD provide useful inductive biases for EHR foundation models, and even lightweight ways of encoding that structure can yield consistent gains over flat code representations. Second, incorporating clinically meaningful structure can also help under distribution shift, and the mechanism of incorporation can impact: graph-based hierarchy encoding transfers more robustly across datasets than token-level hierarchy injection. These insights are relevant beyond ICD-10-CM and can be extended to other structured clinical vocabularies, such as ATC and procedure ontologies, and to other representation learning methods.

\section{Related Work}

\subsection{Sequential and Transformer-based EHR representation learning}
Early work on longitudinal EHR modeling established the importance of learning from temporally ordered patient histories. Doctor AI \citep{choi2016doctor} used recurrent neural networks to predict future clinical events from visit sequences, while RETAIN \citep{choi2016retain} introduced a reverse-time attention mechanism to improve interpretability in healthcare prediction. Dipole \citep{ma2017dipole} further demonstrated that attention-based sequence models can improve diagnosis prediction by capturing dependencies across visits. These methods showed that sequential structure in EHRs is highly informative, but they generally treated diagnosis codes as flat symbols.

More recently, Transformer-based models have become a central approach for structured EHR representation learning. BEHRT \citep{li2020behrt} was an early and influential framework that represents patient histories as sequences of medical codes and applies self-attention to capture dependencies across visits. Med-BERT \citep{rasmy2021medbert} further showed that masked pretraining on large EHR corpora can produce reusable representations for downstream prediction tasks. Subsequent models such as CEHR-BERT \citep{pang2021cehr}, ExBEHRT \citep{rupp2023exbehrt}, and larger-scale foundation models such as Foresight \citep{kraljevic2024foresight} and ETHOS \citep{renc2024zero} have further scaled this paradigm with richer input representations and larger training corpora. 
Despite these advances, Transformer-based EHR models have generally treated diagnosis codes as atomic identifiers.

\subsection{Ontology-aware and Hierarchy-aware Medical Representation Learning}
A parallel line of work has explored how medical ontologies can improve representation learning. Many of these methods were developed in the pre-foundation-model era and are trained end-to-end for a single downstream task rather than as reusable pretrained encoders. GRAM \citep{choi2017gram} incorporates ontology ancestors via attention-weighted aggregation within a recurrent patient encoder, and introduced an intuition central to our study: clinically related diagnoses should share information through their position in a hierarchy. KAME \citep{ma2018kame} extended this with a knowledge-based attention mechanism for diagnosis prediction.
MiME \citep{choi2018mime} explored related ideas by structuring medical codes around clinically meaningful groupings. 
Among ontology-aware methods that use pretraining, G-BERT \citep{shang2019gbert} applies graph-augmented BERT-style pretraining for medication recommendation, though it does not study how different ontology levels contribute.

A complementary line of work has explored purely data-driven approaches to code relationships, learning structure from co-occurrence rather than from an external taxonomy. Med2Vec \citep{choi2016med2vec} demonstrated that skip-gram-style co-occurrence embeddings can recover clinically meaningful code and visit representations without any ontology supervision. Subsequent graph-based methods have modeled empirical dependencies among diagnoses for prediction: \citep{poulain2024graph} builds patient-level heterogeneous graphs that jointly encode diagnoses, procedures, and demographics, learning representations through message passing over clinically grounded relations, while \citep{xi2025breaking} constructs disease co-occurrence graphs from patient records and applies graph neural networks to capture higher-order relational signal beyond pairwise embeddings. 
These approaches recover relational structure directly from data, but unlike ontology-aware methods, they do not leverage existing domain knowledge-based hierarchy.

Across these lines of work, prior methods typically rely on a single source of hierarchical signal. Even when pretrained, they generally do not ablate across hierarchy levels or explicitly examine how hierarchy itself shapes the learned representation.
This leaves open whether hierarchy helps consistently across tasks and architectures, which levels of the ICD hierarchy contribute most, and how hierarchy-aware representations behave under cross-dataset transfer. In contrast, we evaluate two paradigms head-to-head within modern Transformer-based EHR foundation models: \hicdbert\ injects hierarchy at the token-embedding level via string-prefix truncation, requiring no ontology access at training time, while \hicdgraph\ combines ontology-derived hierarchy edges with empirical PMI-weighted co-occurrence edges in a hybrid diagnosis graph, learning hierarchy-aware code embeddings via a graph convolutional network. Across both paradigms, we systematically ablate three hierarchy granularity levels, evaluate on two in-domain tasks and one cross-dataset transfer task, and analyze how hierarchy reshapes the learned embedding geometry, isolating not only \emph{whether} hierarchy helps, but \emph{which} encoding paradigm makes it most useful, \emph{which} levels contribute, and \emph{whether} ontology information is strictly necessary.

\section{Methods}

We represent each patient as a temporally ordered sequence of visits, $\mathcal{V} = (v_1, \dots, v_T)$, where each visit $v_t$ contains one or more ICD diagnosis codes. 
Our goal is to learn patient representations that preserve both temporal context and the hierarchical structure of diagnosis ontologies, and to evaluate whether such structure improves downstream prediction and generalization. 
We study two complementary mechanisms for injecting ICD hierarchy into EHR models, which we collectively refer to as \textbf{H}ierarchical \textbf{ICD} models (\textbf{HICD}). The first is a \emph{token-level} approach, \hicdbert, which extends BEHRT~\citep{li2020behrt} by adding hierarchy-specific embedding tokens for each diagnosis code (Figure~\ref{fig:hicd-comparison}(a)).
The second is a \emph{graph-level} approach, \hicdgraph, which constructs a hierarchy-augmented diagnosis co-occurrence graph, learns code embeddings with a graph convolutional network (GCN)~\citep{kipf2017semi}, and uses those embeddings to initialise a Transformer-based patient encoder (Figure~\ref{fig:hicd-comparison}(b)).
Full implementation details and hyperparameters are provided in Appendix~\ref{app:implementation}.

\subsection{ICD Hierarchy Levels}
\label{sec:hierarchy_levels}

\begin{figure}[t]
    \centering
    \includegraphics[width=0.6\linewidth]{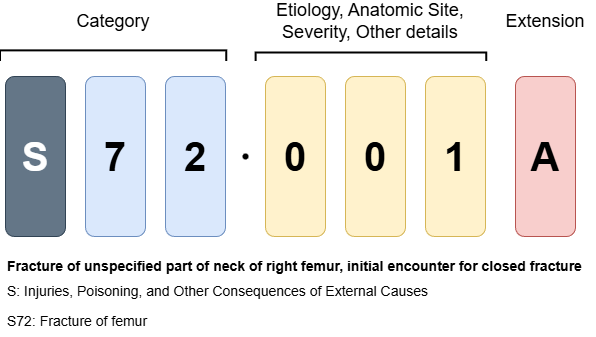}
    \caption{Structure of an ICD-10-CM diagnosis code.}
    \label{fig:icd_code}
\end{figure}

ICD-10-CM codes are organised hierarchically~\citep{who_icd10}. Every code belongs to a \emph{chapter} (a broad body-system or etiology grouping, e.g.\ Chapter~XIX, \emph{Injury, Poisoning and External Causes}), partitioned into officially defined \emph{blocks} of related categories (e.g.\ \texttt{S70--S79}, \emph{Injuries to the hip and thigh}), which contain three-character \emph{categories} (e.g.\ \texttt{S72}, \emph{Fracture of femur}). Characters after the decimal may encode etiology, anatomic site, severity, and encounter details (Fig.~\ref{fig:icd_code}). Each leaf code is nested inside progressively coarser clinical groupings, and this nesting is the inductive bias we inject.

For each diagnosis code, we define three nested hierarchy levels \texttt{G0}, \texttt{G1}, \texttt{G2}, ordered from coarsest to finest. These levels are relative-granularity indices, each architecture operationalises them through its own mechanism (details below). Each ablation setting enables or disables a subset of these levels, yielding all eight combinations $(\texttt{G0},\texttt{G1},\texttt{G2})\in\{0,1\}^3$, including the non-hierarchical baseline $(0,0,0)$, equivalent to BEHRT~\citep{li2020behrt} for \hicdbert\ and to a standard GCN-Transformer without hierarchy edges for \hicdgraph. This setup compares two paradigms: data-driven token encoding, where hierarchy is derived directly from diagnosis code strings, and hybrid graph-based encoding, where hierarchy is injected through ontology-derived graph nodes alongside empirical co-occurrence structure.

\begin{figure}[t]
    \centering
    \subfigure[HICD-BERT][b]{%
        \includegraphics[width=0.32\linewidth]{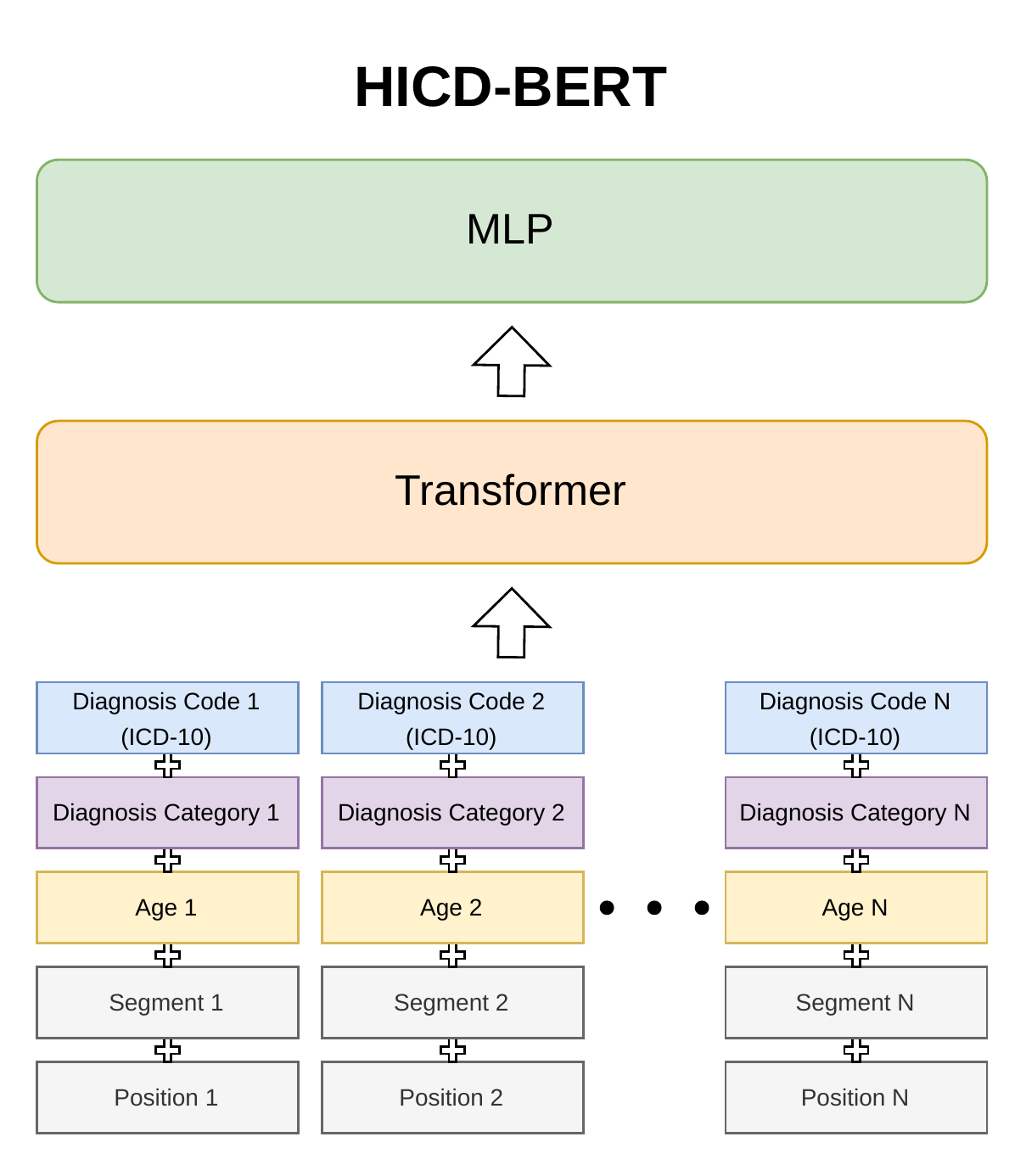}%
        \label{fig:hicd-bert}
    }
    \hfill
    \subfigure[HICD-Graph][b]{%
        \includegraphics[width=0.65\linewidth]{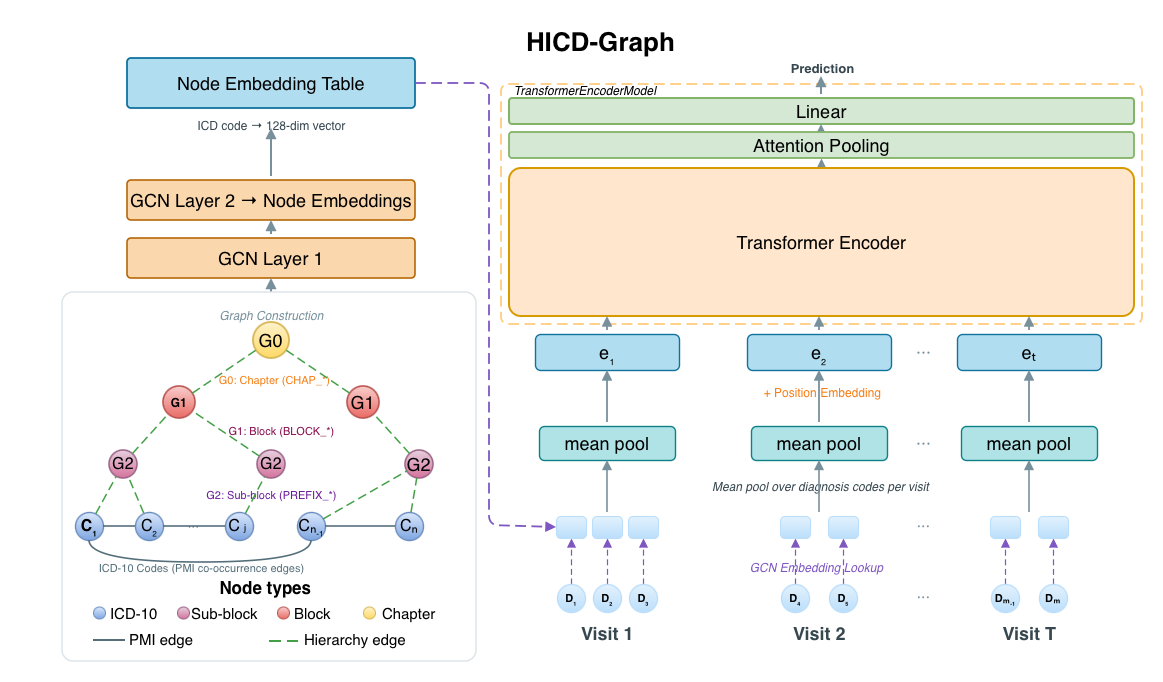}%
        \label{fig:hicd-graph}
    }
    \caption{Comparison of HICD-BERT and HICD-Graph.}
    \label{fig:hicd-comparison}
\end{figure}

\paragraph{\hicdbert\ (data-driven).}
Hierarchy is injected as auxiliary token embeddings using string-prefix truncation (Fig.~\ref{fig:hicd-comparison}(a)), requiring no external ontology at training time. \texttt{G0} is the 1-character alphabetic prefix, \texttt{G1} the 2-character prefix, and \texttt{G2} the 3-character ICD category. These act as data-derived surrogates for increasing hierarchy depth rather than official chapter or block identifiers. For example, \texttt{S72.001A} maps to $(\texttt{G0},\texttt{G1},\texttt{G2})=(\texttt{S},\ \texttt{S7},\ \texttt{S72})$.

\paragraph{\hicdgraph\ (hybrid ontology-based).}
Hierarchy is represented as graph nodes connected to leaf diagnosis codes by weighted edges (Fig.~\ref{fig:hicd-comparison}(b)). These nodes are derived from the ICD-10-CM ontology and are incorporated into a graph that also encodes diagnosis co-occurrence structure. \texttt{G0} denotes the ICD-10 chapter, \texttt{G1} the official ICD-10-CM block, and \texttt{G2} the 2-character prefix group. For example, \texttt{S72.001A} maps to $(\texttt{G0},\texttt{G1},\texttt{G2})=(\text{Chapter~XIX},\ \text{S70--S79},\ \text{S7}x)$. Although \texttt{G1} and \texttt{G2} are closely related, they are not identical: \texttt{G1} follows official ICD-10-CM block boundaries, while \texttt{G2} groups codes by shared first two characters. A worked example of how leaf codes share hierarchy nodes in \hicdgraph\ is provided in Appendix~\ref{app:hicd_graph_example}.

All analyses and comparisons are made \emph{within} each model, and we interpret the levels by relative granularity (coarse to fine) rather than assuming the same label corresponds to identical clinical content across architectures.

\subsection{\hicdbert: Hierarchy-aware BEHRT}
\label{sec:hicdbert}
\hicdbert\ extends BEHRT~\citep{li2020behrt}, a Transformer encoder for longitudinal EHR sequences. In BEHRT, each diagnosis token is represented by the sum of the token, age, positional, and segment embeddings. We augment this
representation with up to three hierarchy-specific embeddings:
\begin{equation}
\mathbf{h}^{(0)}_i =
\mathbf{e}_{\text{code}}(c_i) +
\mathbf{e}_{\text{age}}(a_i) +
\mathbf{e}_{\text{pos}}(p_i) +
\mathbf{e}_{\text{seg}}(s_i) +
\mathbb{I}[G0]\mathbf{e}_{G0}(g_i^{(0)}) +
\mathbb{I}[G1]\mathbf{e}_{G1}(g_i^{(1)}) +
\mathbb{I}[G2]\mathbf{e}_{G2}(g_i^{(2)}),
\end{equation}
where $c_i$ is the diagnosis token, $a_i$ is age, $p_i$ is position, $s_i$ is the segment identifier, and $g_i^{(0)}, g_i^{(1)}, g_i^{(2)}$ are the group identifiers at hierarchy levels $G0$, $G1$, and $G2$.
This design preserves the original BEHRT tokenization while directly injecting the coarse-to-fine ICD structure into the embedding layer.
The resulting sequence is processed by a stack of Transformer encoder blocks~\citep{vaswani2017attention}. 
Each block contains multi-head self-attention, a position-wise feed-forward network, residual connections, and layer normalization. 
The hierarchy-aware embedding layer is the only change to the BEHRT backbone. 
Additional architectural and training details are provided in Appendix~\ref{app:hicdbert_details}.

We pretrain \hicdbert\ using a masked language modeling (MLM) objective over the
diagnosis sequence. Given a masked sequence $\tilde{\mathbf{x}}$, the model
predicts the original diagnosis token at masked positions:
\begin{equation}
\mathcal{L}_{\text{MLM}} =
- \sum_{i \in \mathcal{M}} \log p(c_i \mid \tilde{\mathbf{x}}),
\end{equation}
where $\mathcal{M}$ denotes the set of masked positions.

For downstream prediction, we initialize the encoder from the pretrained MLM checkpoint and attach a task-specific multilayer perceptron to the final \texttt{[CLS]} representation. Binary tasks are optimized with binary cross-entropy with logits, while multiclass tasks are optimized with
cross-entropy.

\subsection{\hicdgraph: Graph-based Hierarchy Modeling}
\label{sec:hicdgraph}

Our second model injects ICD hierarchy at the code-graph level rather than the token-embedding level.
We first construct a diagnosis co-occurrence graph over ICD codes, where horizontal edges reflect empirical co-occurrence within patient visits
and are weighted by pointwise mutual information (PMI):
\begin{equation}
\text{PMI}(a,b) = \log \frac{P(a,b)}{P(a)P(b)} + \epsilon,
\end{equation}
where $P(a)$ and $P(b)$ are marginal code frequencies and $P(a,b)$ is their joint co-occurrence probability.
We retain only positive PMI edges above a percentile threshold to suppress weak associations.
We augment this empirical graph with uniform-weight ICD-10 hierarchy edges linking each diagnosis node to up to three enabled ancestor levels, \texttt{G2}, \texttt{G1}, and \texttt{G0} (Section~\ref{sec:hierarchy_levels}, Fig. \ref{fig:hicd_graph_example}), yielding a hybrid graph construction that combines data-driven PMI edges with ontological hierarchy edges.
Given the resulting graph, we learn code embeddings using a two-layer GCN~\citep{kipf2017semi} trained with a link-prediction objective:
\begin{equation}
\mathcal{L}_{\text{graph}} =
-\sum_{(u,v)\in\mathcal{E}} \log \sigma(\mathbf{z}_u^\top \mathbf{z}_v)
-\sum_{(u,v^-)\notin\mathcal{E}} \log \left(1-\sigma(\mathbf{z}_u^\top \mathbf{z}_{v^-})\right),
\end{equation}
where $\mathbf{z}_u$ and $\mathbf{z}_v$ denote node embeddings and $\mathcal{E}$ is the set of graph edges.

The learned code embeddings are then used to initialise the embedding table of a patient-level Transformer encoder.
For each visit, we compute a visit-level representation by applying masked mean pooling over the embeddings of the diagnosis codes (and, when enabled, their hierarchy ancestor codes) observed in that visit.
The temporally ordered sequence of visit representations is then processed by a Transformer encoder, followed by attention-based pooling across visits to produce a patient-level representation, which is passed to a task-specific classification head.
Additional implementation details are provided in Appendix~\ref{app:graph_details}.

\subsection{Experimental Setup}
\subsubsection{Datasets}
\label{subsec:datasets}

\paragraph{MIMIC-IV} 
We use data from the Medical Information Mart for Intensive Care IV
(MIMIC\mbox{-}IV), a large, de\mbox{-}identified clinical database developed at
Beth Israel Deaconess Medical Center in Boston, Massachusetts
\citep{johnson_mimic-iv_2023}. MIMIC\mbox{-}IV contains electronic health
records for hospital admissions between 2008 and 2019.
For this work, we restrict our use of MIMIC\mbox{-}IV to
\emph{diagnosis information} from the hospital-wide (\texttt{hosp}) module.
This module covers all inpatient admissions, not only ICU stays, and stores
structured diagnosis codes assigned to each hospital encounter.
Diagnoses are encoded using International Classification of Diseases,
Ninth Revision, Clinical Modification (ICD\mbox{-}9\mbox{-}CM)(mapped to ICD-10 in our work) and Tenth
Revision, Clinical Modification (ICD\mbox{-}10\mbox{-}CM). Each diagnosis code
is linked to dictionary tables providing a textual description and hierarchical
groupings, which we leverage to define clinical conditions and construct label spaces.

\paragraph{eICU Collaborative Research Database}
The eICU Collaborative Research Database (eICU-CRD) is a multi-center critical
care database containing de-identified health data for over 200{,}000 ICU
admissions from more than 200 hospitals across the United States
\citep{Pollard2018}. Data were collected through the Philips eICU telehealth
program and span a wide range of ICU types, including medical, surgical,
cardiac, and mixed units. eICU-CRD provides diagnosis codes recorded using
ICD\mbox{-}9\mbox{-}CM and ICD\mbox{-}10\mbox{-}CM, as well as proprietary
diagnosis fields used within the telehealth platform.

Both MIMIC-IV and eICU-CRD are de-identified, hosted on PhysioNet~\citep{goldberger2000physiobank}, and available to credentialed researchers who complete the required data use training and agreements.

\subsubsection{Task Definitions} 

We evaluate all models on two binary classification tasks defined over
patient visit sequences from MIMIC-IV, and one binary classification task
on eICU for transfer evaluation. For all MIMIC-IV tasks, the final visit of
each patient is excluded from the prediction set, as no subsequent outcome
label can be observed.

\paragraph{30-Day Readmission (MIMIC-IV).}
Given a patient's visit history up to and including visit $v_t$, the model
predicts whether the patient will be readmitted within 30 days of discharge.
The binary label is defined as
\begin{equation}
    y_t^{\text{readmit}} =
    \mathbb{I}\!\left[
        \mathrm{admittime}(v_{t+1}) - \mathrm{dischtime}(v_t) < 30 \text{ days}
    \right].
\end{equation}
This task serves as a standard clinical benchmark for assessing whether
hierarchy-aware representations capture patient risk signals that are
predictive of near-term acute care utilization.

\paragraph{ Emergency-Admission Prediction (MIMIC-IV).}
Given a patient's visit history up to and including visit $v_t$, the model
predicts whether the immediately following visit $v_{t+1}$ is an emergency
admission. MIMIC-IV records nine admission types; we define emergency
admissions as the union of \texttt{URGENT}, \texttt{EW EMER.},
\texttt{EU OBSERVATION}, \texttt{OBSERVATION ADMIT},
\texttt{AMBULATORY OBSERVATION}, \texttt{DIRECT EMER.}, and
\texttt{DIRECT OBSERVATION}, based on the visit type information provided by
MIMIC-IV. The binary label is
\begin{equation}
    y_t^{\text{type}} =
    \mathbb{I}\!\left[
        \mathrm{admission\_type}(v_{t+1}) \in \mathcal{T}_{\text{EM}}
    \right],
\end{equation}
where $\mathcal{T}_{\text{EM}}$ denotes the set of emergency admission types
defined above.
This task evaluates whether the learned representations encode temporal
diagnostic patterns predictive of the urgency of future care.

\paragraph{ICU Readmission Prediction (eICU).}
The eICU Collaborative Research Database organizes data at the level of
individual ICU unit stays, with multiple unit stays possible within a single
hospital admission, linked by a common
\texttt{patientHealthSystemStayID}. Given the diagnosis codes observed during
ICU unit stay $u_k$ for a patient, the model predicts whether the patient will
be readmitted to the ICU within the same hospital admission. The binary label
is
\begin{equation}
    y_k^{\text{readmit}} =
    \mathbb{I}\!\left[
        u_k \text{ is not the final ICU unit stay within the hospital admission}
    \right].
\end{equation}
This task is used exclusively for transfer evaluation: models are pretrained
on MIMIC-IV and only a classifier head is fine-tuned on eICU, with the encoder
kept frozen.

We split the data into an 80:10:10 ratio to create a train/validation/test split shared across
all hierarchy ablations for a model. For \hicdbert, we first pretrain the model with
masked language modeling (ref. Sec. \ref{sec:hicdbert}) and then fine-tune it for each downstream task. For
\hicdgraph, we first learn hierarchy-aware code embeddings from the diagnosis
graph and then use these embeddings to initialize the downstream
Transformer-based patient encoder. In \hicdbert, we evaluate all subsets of
the three hierarchy levels \texttt{G0}, \texttt{G1}, and \texttt{G2}, which
correspond to enabling or disabling hierarchy embeddings; in \hicdgraph, they
correspond to enabling or disabling the associated hierarchy edges and ancestor
nodes.

We evaluate performance using  F1, and AUROC scores. For
transfer experiments, models trained on MIMIC-IV are evaluated on eICU without
modifying the hierarchy construction pipeline. Exact optimization settings and
hyperparameters are provided in Appendix~\ref{app:implementation}.

\section{Results}
We report the in-domain downstream performance of hierarchy-aware modeling, cross-dataset transfer under a frozen-encoder probe setting, and embedding-level analyses.
We evaluate all eight combinations of the three ICD-10 hierarchy levels (\texttt{G0}, \texttt{G1}, \texttt{G2}) across both \hicdbert\ and \hicdgraph, resulting in \texttt{28} pairwise comparisons using \emph{2 models $\times$ 2 tasks} against the no-hierarchy baseline within each paradigm: BEHRT \citep{li2020behrt} for \hicdbert\ and a GCN + Transformer encoder for \hicdgraph.
We report statistical significance using the paired DeLong test~\citep{delong1988comparing} for all pairwise comparisons of binary AUROC on the same test set.
For F1-macro, we use a paired bootstrap over 1{,}000 test-set resamples and a two-sided p-value, following prior work \citep{berg2012empirical}.
\begin{table*}[t]
\centering
\small
\setlength{\tabcolsep}{6pt}
\caption{30-day readmission results for \hicdbert and \hicdgraph across hierarchy ablations. \textsuperscript{*}\,$p<0.05$ from pairwise DeLong tests on AUROC and paired bootstrap tests on F1-macro, both against the no-hierarchy baseline.}
\label{tab:readmit_side_by_side}
\begin{tabular}{llcc|cc}
\toprule
Group & Hierarchy encoding & \multicolumn{2}{c|}{\hicdbert} & \multicolumn{2}{c}{\hicdgraph} \\
\cmidrule(lr){3-4} \cmidrule(l){5-6}
& & F1-macro & AUROC & F1-macro & AUROC \\
\midrule
\cellcolor{gray!12} Baseline & \cellcolor{gray!12} No hierarchy & \cellcolor{gray!12} 0.5786 & \cellcolor{gray!12} 0.6808 & \cellcolor{gray!12} 0.5976 & \cellcolor{gray!12} 0.6992 \\
\midrule
& G0    & 0.5825 & 0.6858\textsuperscript{*} & 0.6025\textsuperscript{*} & 0.7086\textsuperscript{*} \\
\cellcolor{blue!8} Single-level & G1    & 0.5740\textsuperscript{*}  & 0.6811 & 0.6006\textsuperscript{*} & 0.7079\textsuperscript{*} \\
 & G2    & 0.5896\textsuperscript{*}  & 0.6921\textsuperscript{*} & 0.6005\textsuperscript{*} & 0.7086\textsuperscript{*} \\
\midrule
 & G0+G1 & 0.5696\textsuperscript{*}  & 0.6863\textsuperscript{*} & \textbf{0.6088}\textsuperscript{*} & 0.7105\textsuperscript{*} \\
\cellcolor{red!8} Two-Level & G0+G2 & \textbf{0.5953}\textsuperscript{*}  & 0.6918\textsuperscript{*} & 0.5999 & 0.7097\textsuperscript{*} \\
 & G1+G2 & 0.5834\textsuperscript{*}  & 0.6917\textsuperscript{*} & 0.6000 & 0.7113\textsuperscript{*} \\
\midrule
\cellcolor{orange!10}All-Levels & G0+G1+G2 & 0.5731\textsuperscript{*}  & \textbf{0.6923}\textsuperscript{*} & 0.5991 & \textbf{0.7114}\textsuperscript{*} \\
\bottomrule
\end{tabular}
\end{table*}

\subsection{30-day Readmission Task}
Table~\ref{tab:readmit_side_by_side} shows that adding hierarchical structure improves 30-day readmission prediction for both \hicdbert\ and \hicdgraph\ relative to the no-hierarchy baseline (DeLong $p<0.05$ for 6 of 7 \hicdbert\ variants and all 7 \hicdgraph\ variants).

\vspace{0.5em}
\noindent\textbf{For \hicdbert, multi-level hierarchy outperforms single-level, with the finest level driving the gains.}
The strongest F1-macro is achieved by the two-level \texttt{G0{+}G2} configuration (0.5953), while the best AUROC is obtained when all hierarchy levels are enabled (0.6923). Among single-level configurations, the finest level \texttt{G2} yields the strongest improvement (AUROC 0.6921), while the mid-level \texttt{G1} is the only configuration that does not significantly improve over baseline, though it contributes meaningfully when combined with other levels. This suggests that for token-level hierarchy injection, fine-grained groupings are the primary driver of performance gains, and that mid-level groupings serve as connective structure between coarser and finer hierarchy rather than as standalone signal.

\vspace{0.5em}
\noindent\textbf{For \hicdgraph, more hierarchy levels yield progressively better readmission prediction.}
All seven hierarchy configurations yield statistically significant improvements over the no-hierarchy baseline, and a clear monotonic pattern emerges: single-level configurations improve AUROC by $1.3\%$ on average over the baseline, two-level configurations by $1.6\%$, and the all-levels \texttt{G0{+}G1{+}G2} configuration by $1.7\%$. This progressive improvement suggests that each additional hierarchy level contributes complementary structure to the diagnosis graph, and that the graph-based architecture is able to effectively integrate information from multiple granularity levels simultaneously.

\vspace{0.5em}
Taken together, the consistency of hierarchy benefits across two different pretraining objectives, a Transformer-based encoder pretrained with masked language modeling (\hicdbert) and a GCN-initialized Transformer pretrained with link prediction on a diagnosis co-occurrence graph (\hicdgraph), supports that ICD hierarchy is a useful inductive bias for readmission prediction.
While the absolute AUROC improvements are modest, they are statistically robust and architecturally consistent; we discuss the practical significance of these effect sizes in Section~\ref{sec:discussion}.

%%%%%%%%%%%

\begin{table*}[t]
\centering
\small
\setlength{\tabcolsep}{6pt}
\caption{Emergency admission prediction results for \hicdbert and \hicdgraph across hierarchy ablations.\textsuperscript{*}\,$p<0.05$ from pairwise DeLong tests on AUROC and paired bootstrap tests on F1-macro, both against the no-hierarchy baseline.}
\label{tab:emvisit_side_by_side}
\begin{tabular}{llcc|cc}
\toprule
Group & Hierarchy encoding & \multicolumn{2}{c|}{\hicdbert} & \multicolumn{2}{c}{\hicdgraph} \\
\cmidrule(lr){3-4} \cmidrule(l){5-6}
& & F1-macro & AUROC & F1-macro & AUROC \\
\midrule
\cellcolor{gray!12}Baseline & \cellcolor{gray!12} No hierarchy & \cellcolor{gray!12} 0.4945 & \cellcolor{gray!12} 0.7682 & \cellcolor{gray!12} 0.6083 & \cellcolor{gray!12} 0.7495 \\
\midrule
& G0 & 0.4846\textsuperscript{*} & 0.7848\textsuperscript{*} & 0.6373\textsuperscript{*} & 0.7538\textsuperscript{*} \\
\cellcolor{blue!8} Single-level & G1 & 0.5107\textsuperscript{*} & 0.7698 & 0.6209\textsuperscript{*} & 0.7555\textsuperscript{*} \\
& G2 & \textbf{0.5200}\textsuperscript{*} & \textbf{0.7941}\textsuperscript{*} & 0.6269\textsuperscript{*} & 0.7556\textsuperscript{*} \\
\midrule
& G0+G1 & 0.5138\textsuperscript{*} & 0.7838\textsuperscript{*} & 0.6343\textsuperscript{*} & 0.7557\textsuperscript{*} \\
\cellcolor{red!8} Two-Level & G0+G2 & 0.5159\textsuperscript{*} & 0.7924\textsuperscript{*} & 0.6314\textsuperscript{*} & 0.7557\textsuperscript{*} \\
& G1+G2 & 0.5118 \textsuperscript{*}& 0.7903\textsuperscript{*} & 0.6313\textsuperscript{*} & \textbf{0.7577}\textsuperscript{*} \\
\midrule
\cellcolor{orange!10}All-Levels & G0+G1+G2 & 0.4988 & 0.7917\textsuperscript{*} & 0.6339\textsuperscript{*}  & 0.7567\textsuperscript{*} \\
\bottomrule
\end{tabular}
\end{table*}

\subsection{Emergency Admission Prediction } 
We report the results for emergency admission prediction in Table~\ref{tab:emvisit_side_by_side}.
Here again, we note that hierarchy improves over the no-hierarchy baseline, but the pattern of which configurations perform best differs from the readmission task.

\vspace{0.5em}
\noindent\textbf{For \hicdbert, a single hierarchy level appears to be as effective as the full hierarchy.}
Unlike the readmission setting, where the all-levels configuration achieves the best AUROC, the single-level \texttt{G2} model performs as strongly as the all-levels variant on emergency-visit prediction, with both achieving comparable AUROC. As in readmission, the mid-level \texttt{G1} alone does not significantly improve over baseline, though it does yield a significant F1-macro gain. This task-dependent shift suggests that emergency-visit prediction can drive performance from a specific level of abstraction, and that adding coarser hierarchy levels on top of the finest grouping provides no additional benefit for this task.

\vspace{0.5em}
\noindent\textbf{For \hicdgraph, the monotonic trend from readmission holds.}
All seven variants significantly outperform the no-hierarchy baseline (all $p<0.001$), and as in readmission, AUROC improves progressively from single-level to two-level to all-levels configurations. The best individual configuration is the two-level \texttt{G1{+}G2}, which achieves comparable or slightly better AUROC than the all-levels variant. This suggests that for emergency-visit prediction, the combination of intermediate and fine-grained hierarchy captures the most relevant diagnostic structure, and that adding the coarsest level (ICD chapter) on top provides no further benefit for this task.

\vspace{0.5em}
Across both in-domain tasks, hierarchy-aware models outperform the no-hierarchy baseline in 26 of 28 pairwise  AUROC  comparisons, establishing ICD hierarchy as a \textbf{robustly effective inductive bias} regardless of task or architecture. Our systematic ablation further reveals that while hierarchy consistently improves over baseline, the optimal configuration varies by task, reinforcing the value of treating hierarchy levels as controllable design choices.

\begin{table*}[t]
\centering
\small
\setlength{\tabcolsep}{6pt}
\caption{\textsc{Cross Dataset Transfer:} Probe results on eICU readmission task for \hicdbert and \hicdgraph across hierarchy ablations. \textsuperscript{*}\,$p<0.05$ from pairwise DeLong tests on AUROC and paired bootstrap tests on F1-macro, both against the no-hierarchy baseline.}
\label{tab:eicu_zeroshot_transfer}
\begin{tabular}{llcc|cc}
\toprule
Group & Hierarchy encoding & \multicolumn{2}{c|}{\hicdbert} & \multicolumn{2}{c}{\hicdgraph} \\
\cmidrule(lr){3-4} \cmidrule(l){5-6}
& & F1-macro & AUROC & F1-macro & AUROC \\
\midrule
\cellcolor{gray!12}Baseline & \cellcolor{gray!12} No hierarchy & \cellcolor{gray!12} 0.455 & \cellcolor{gray!12} 0.565 & \cellcolor{gray!12} 0.380\textsuperscript{*} & \cellcolor{gray!12} 0.565 \\
\midrule
 & G0 & 0.453 & \textbf{0.571}\textsuperscript{*} & 0.396\textsuperscript{*} & 0.576\textsuperscript{*} \\
\cellcolor{blue!8} Single-level & G1 & 0.447\textsuperscript{*} & 0.564 & 0.487\textsuperscript{*} & 0.578\textsuperscript{*} \\
& G2 & \textbf{0.485}\textsuperscript{*} & 0.565 & 0.461\textsuperscript{*} & 0.583\textsuperscript{*} \\
\midrule
& G0+G1 & 0.432\textsuperscript{*} & 0.568 & 0.467\textsuperscript{*} & 0.582\textsuperscript{*} \\
\cellcolor{red!8} Two-Level & G0+G2 & 0.433\textsuperscript{*} & 0.567 & \textbf{0.487\textsuperscript{*}} & 0.579\textsuperscript{*} \\
& G1+G2 & 0.475\textsuperscript{*} & 0.568 & 0.453\textsuperscript{*} & 0.574 \\
\midrule
\cellcolor{orange!10}All-Levels & G0+G1+G2 & 0.474\textsuperscript{*} & 0.565 & 0.474\textsuperscript{*} & \textbf{0.587}\textsuperscript{*} \\
\bottomrule
\end{tabular}
\end{table*}

\subsection{Cross-Dataset Transfer Results}
Table~\ref{tab:eicu_zeroshot_transfer} presents cross-dataset transfer results on eICU under a frozen-encoder probe protocol. 
In this setting, the encoder pretrained on MIMIC-IV is kept frozen, and only a task-specific classifier head is trained using eICU ICU-readmission labels.
This evaluation measures whether the hierarchy-aware representations learned on one institution's data capture structure that generalises to a different patient population.
We note that both \hicdbert\ and \hicdgraph\ retain meaningful predictive signal under this setting, with all configurations outperforming chance-level prediction (AUROC $> 0.55$). However, the pattern of which hierarchy configurations benefit transfer differs between the two architectures and from the in-domain results.

\vspace{0.5em}
\noindent\textbf{Hierarchy benefits transfer more in \hicdgraph\ than in \hicdbert.} 
For \hicdgraph, 6 of 7 hierarchy variants significantly improve over the no-hierarchy baseline, with the all-levels \texttt{G0{+}G1{+}G2} configuration achieving the strongest AUROC (0.587). This demonstrates that the graph-based hierarchy encoding produces representations whose structure transfers robustly across datasets.
In contrast, for \hicdbert, only the coarsest single-level configuration \texttt{G0} improves AUROC over baseline (0.571); the remaining 6 variants are statistically indistinguishable from the no-hierarchy baseline, and F1-macro results are mixed, with only some variants improving F1 (\texttt{G2}, \texttt{G12}, \texttt{G0{+}G1{+}G2}).
. This is notable because \hicdbert\ is the stronger beneficiary of hierarchy in the in-domain setting, where 6 of 7 variants achieve improvements, yet these gains do not survive the distributional shift to eICU.
The asymmetry may reflect a difference in what the two encoders learn: the co-occurrence graph underlying \hicdgraph\ captures inter-code relationships that can be more institution-invariant, whereas the sequential token patterns learned by \hicdbert\ are more tightly coupled to MIMIC-IV visit structures. Consistent with this interpretation, only the coarsest level \texttt{G0} transfers for \hicdbert\ as coarse groupings can be less susceptible to dataset-specific coding variation than finer-grained levels.

\vspace{0.5em}
\noindent\textbf{Hierarchy aware pretrained representations capture transferable diagnosis structure.}
Despite the modest absolute AUROC values (0.56--0.59), these results demonstrate that hierarchy-augmented pretraining produces more robust representations that can transfer well across datasets.
The frozen encoder has not seen eICU patients or diagnoses during pretraining, yet hierarchy-aware configurations, particularly in \hicdgraph, achieve statistically significant improvements over the no-hierarchy baseline.
This supports that the ICD hierarchy provides an inductive bias that is not merely corpus-specific but reflects underlying relationships that can transfer well across datasets.

\subsection{Embedding Analysis}

\begin{figure}[t]
    \centering
    \subfigure[Pairwise similarity.][t]{%
        \includegraphics[width=0.46\linewidth]{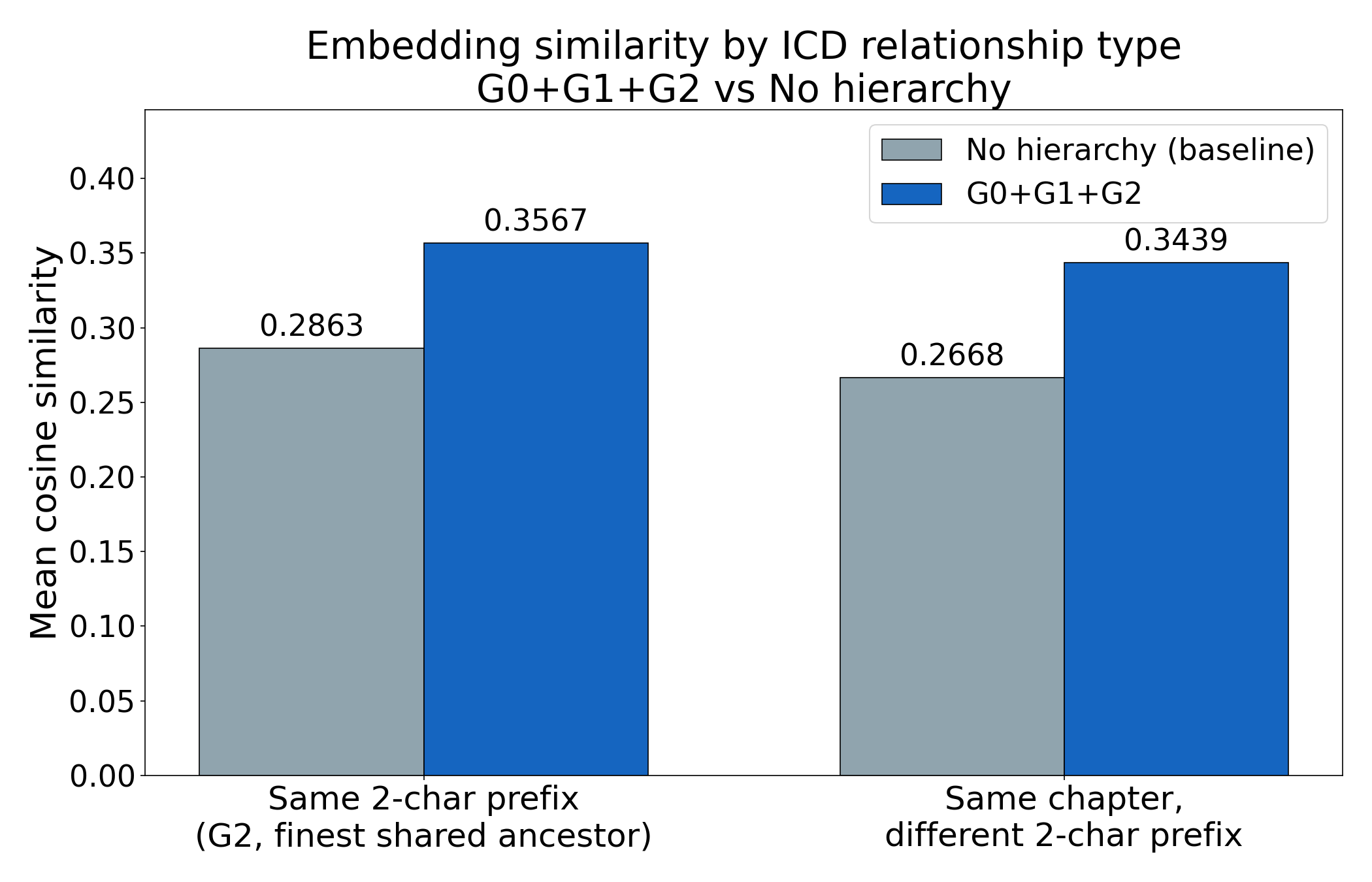}%
        \label{fig:embedding_analysis_A}
    }
    \hfill
    \subfigure[Within-group similarity.][t]{%
        \includegraphics[width=0.48\linewidth]{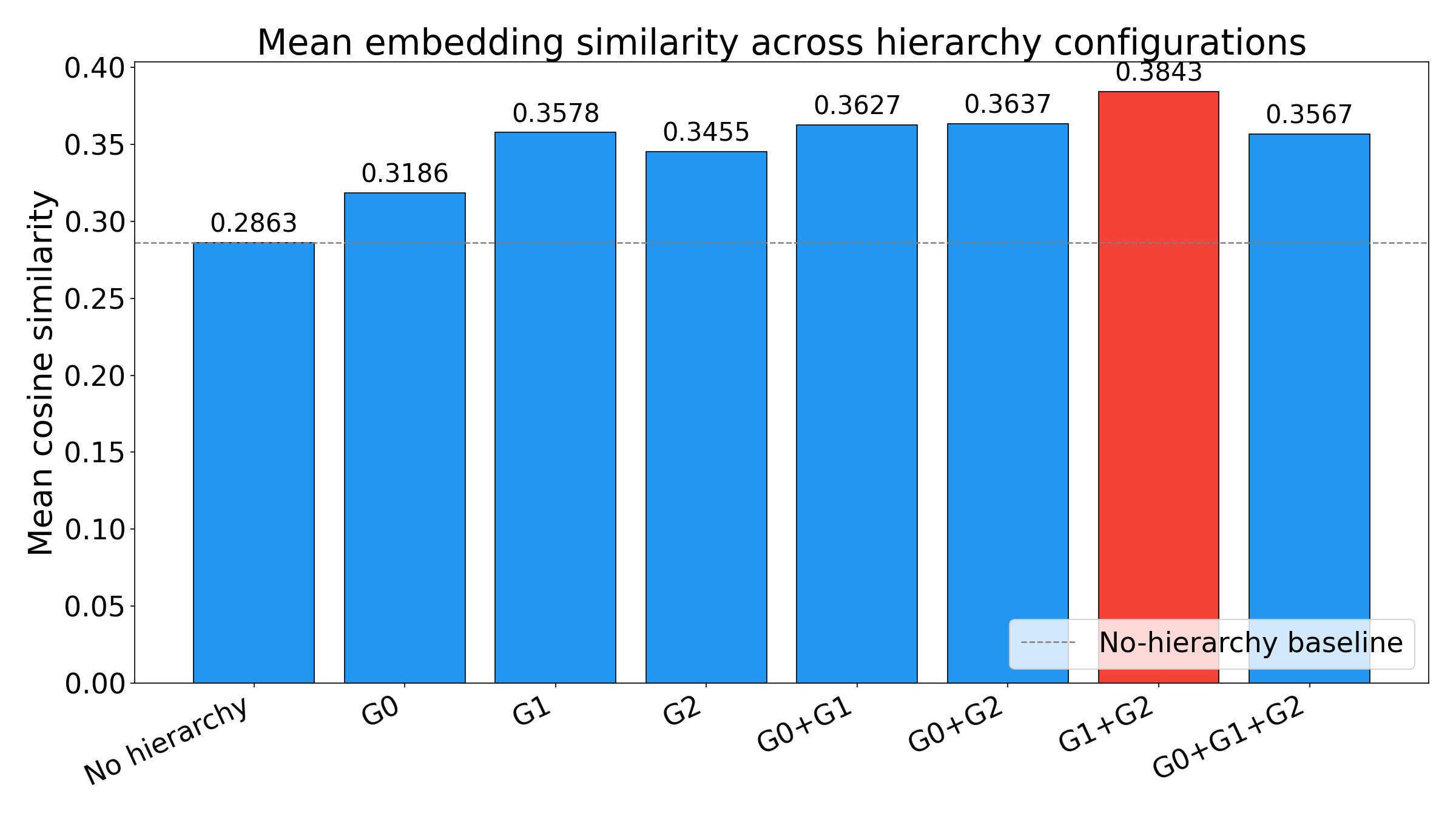}%
        \label{fig:embedding_analysis_B}
    }
    \caption{Embedding analysis for \hicdgraph. All hierarchy-aware variants increase semantic clustering over the baseline, with \texttt{G1{+}G2} achieving the strongest local clustering.}

    \label{fig:embedding_analysis}
\end{figure}

We analyze the learned GCN node embeddings in \hicdgraph\ for leaf ICD-10 diagnosis codes to assess how hierarchical supervision changes the structure of the diagnosis-code representation space.
We study pairwise cosine similarity under two ICD relationship types, defined by the shared ancestor in the ICD-10 ontology: codes sharing the same 2-character prefix (e.g.\ all codes beginning with \texttt{S7}), which corresponds to the model's \texttt{G2} grouping level, and codes from the same ICD-10 chapter but with different 2-character prefixes.

\vspace{0.5em}
\noindent\textbf{Hierarchy strengthens clinically meaningful similarity structure.}
Figure~\ref{fig:embedding_analysis}(a) compares the mean cosine similarity across these two relationship types for the no-hierarchy baseline and the \texttt{G0{+}G1{+}G2} configuration.
Relative to the baseline, the \texttt{G0{+}G1{+}G2} configuration increases similarity in both groups, from 0.286 to 0.357 among codes sharing the same 2-character prefix, and from 0.267 to 0.344 among codes within the same chapter but with different prefixes.
The resulting embedding space preserves a meaningful similarity gradient: codes sharing the finest ancestor (\texttt{G2}) are more similar than broader within-chapter codes, supporting that hierarchical supervision strengthens the clinically relevant organization of the embedding space.

\vspace{0.5em}
\noindent\textbf{All hierarchy configurations improve local code similarity.}
Figure~\ref{fig:embedding_analysis}(b) examines how each hierarchy configuration affects pairwise similarity among diagnosis codes.
All hierarchy-aware variants improve over the no-hierarchy baseline (0.286), confirming that explicit hierarchical information consistently pulls related diagnoses closer together in embedding space.
The strongest effect is obtained when the mid-level and fine-level hierarchies are combined (\texttt{G1{+}G2}, 0.384), indicating that these two levels together capture the most clinically relevant grouping structure for organising related diagnosis codes in embedding space.

\vspace{0.5em}
\noindent\textbf{Embedding structure predicts downstream performance.}
Notably, these embedding-level patterns are also predictive of downstream task performance. On readmission prediction (Table~\ref{tab:readmit_side_by_side}), the \texttt{G1{+}G2} configuration, which produces the tightest local code clusters in embedding space, also achieves among the strongest AUROC for \hicdgraph, while the coarsest level \texttt{G0}, which yields the smallest embedding similarity gains, delivers the weakest single-level task performance.
This alignment between representation geometry and predictive accuracy suggests that hierarchy improves \hicdgraph\ performance by shaping the embedding space in clinically meaningful ways that benefit downstream discrimination.

\vspace{0.5em}
\noindent\textbf{Hierarchy effects are heterogeneous across chapters in our pretraining data}
Figure~\ref{fig:intra_chapter_cosine} (Appendix) shows the intra-chapter cosine similarity for each ICD-10 chapter across all eight hierarchy configurations on our pretraining data, MIMIC-IV. 
Across the majority of chapters in our pretraining data, cosine similarity increases with the addition of hierarchy, confirming the global trend observed in Figure~\ref{fig:embedding_analysis}. 

However, the magnitude of improvement is heterogeneous: chapters with higher baseline similarity, such as Chapter~XVIII (0.48 $\rightarrow$ 0.61, +27\%) and Chapter~XXII (0.70 $\rightarrow$ 0.90, +29\%), exhibit comparatively modest relative gains, while lower-baseline chapters such as Chapter~XVI (0.09 $\rightarrow$ 0.26, +189\%) and Chapter~XIX (0.05 $\rightarrow$ 0.10, +100\%) show substantially larger proportional improvements under hierarchy. This suggests that hierarchical supervision is slightly more beneficial in chapters with greater code diversity, where the representation space is less coherent at baseline, and the model can be optimized more effectively by incorporating ontological structure.

\section{Discussion}
\label{sec:discussion}

This work investigates ICD hierarchy as an inductive bias for EHR representation learning, evaluating two complementary mechanisms, token-level injection (\hicdbert) and graph-level encoding (\hicdgraph).
Our results support the view that the hierarchical structure already present in ICD coding systems is an \textit{underutilised signal} in EHR foundation models, and that modeling it explicitly benefits downstream performance.

\paragraph{Key Findings.}
Across both in-domain tasks, the finest hierarchy level \texttt{G2} is consistently the strongest single-level configuration for \hicdbert, while \hicdgraph\ benefits progressively from adding more hierarchy levels. For \hicdbert, the mid-level \texttt{G1} provides no significant improvement in isolation on either task, though it contributes when combined with other levels, suggesting that mid-level groupings serve as connective structure rather than a standalone signal. The optimal hierarchy depth is also task-dependent: readmission benefits from combining all levels, while emergency-visit prediction achieves comparable performance from a single fine-grained level alone. 
On cross-dataset transfer, hierarchy benefits \hicdgraph\ robustly (6 of 7 variants significant) but largely vanish for \hicdbert, where only the coarsest level \texttt{G0} transfers. This suggests that graph-based relational structure, anchored in the ICD ontology, is more dataset-invariant than sequential token patterns. The embedding analysis provides mechanistic support: hierarchy constrains the embedding space to respect clinical relationships, and the configurations producing the tightest local code clusters (\texttt{G1{+}G2}) also achieve among the strongest downstream AUROC, confirming that the gains arise from meaningful representational structure.

\paragraph{Practical significance of effect sizes.}
Readmission and emergency-visit prediction are inherently difficult, multifactorial outcomes that depend on many factors beyond \textit{diagnosis codes alone}. Against this backdrop, the AUROC improvements from hierarchy are statistically robust (26 of 28 comparisons significant) and consistent across two models and two tasks, supporting hierarchy as a general inductive bias. Importantly, in our setup, we incorporated hierarchy using only minimal modifications to existing architectures: additive token embeddings for \hicdbert\ and additional graph edges for \hicdgraph\, with no changes to the core encoder, pretraining objective, or training pipeline. This \textit{intentional low cost integration}, combined with the consistent improvements observed, suggests that the ICD hierarchy should be included as a strong prior in EHR representation learning.

\paragraph{Limitations.}
Our evaluation covers two binary clinical tasks on MIMIC-IV and one transfer task on eICU; the generalizability of our findings to other clinical outcomes, institutions, and EHR systems remains to be validated. Our analysis is restricted to ICD diagnosis codes; whether the observed benefits extend to other coded data in EHR foundational models remains an open question. The hierarchy levels (\texttt{G0}, \texttt{G1}, \texttt{G2}) are intentionally defined differently across \hicdgraph and \hicdbert to evaluate two hierarchy construction strategies: ontology-derived grouping versus lightweight data-driven prefix grouping. All comparisons are therefore made within each model, while future work could analyze other encodings across architectures to isolate the effect of the hierarchy definition itself.
Our hierarchy experiments are limited to a fixed set of manually defined levels, so we do not test whether the optimal hierarchy depth could be learned automatically for a given task or model.

\paragraph{Future Work.}
A natural next step is to extend this framework to other structured clinical vocabularies with analogous ontological structure, such as ATC codes for medications, which organize drugs into therapeutic, pharmacological, and chemical subgroups across multiple levels. We also plan to explore deeper integration mechanisms that go beyond the additive approaches studied here, including hierarchy-aware pretraining objectives that explicitly optimize for ontological coherence, adaptive mechanisms that learn to weight or select different hierarchy levels during training, and multi-ontology fusion that jointly encodes diagnosis, medication, and procedure hierarchies within a single model. We believe such approaches could further amplify the practical gains observed in this work.

\newpage
\bibliography{references}

@article{johnson_mimic-iv_2023,
	title = {{MIMIC}-{IV}, a freely accessible electronic health record dataset},
	volume = {10},
	issn = {2052-4463},
	url = {https://doi.org/10.1038/s41597-022-01899-x},
	doi = {10.1038/s41597-022-01899-x},
	number = {1},
	journal = {Scientific Data},
	author = {Johnson, Alistair E. W. and Bulgarelli, Lucas and Shen, Lu and Gayles, Alvin and Shammout, Ayad and Horng, Steven and Pollard, Tom J. and Hao, Sicheng and Moody, Benjamin and Gow, Brian and Lehman, Li-wei H. and Celi, Leo A. and Mark, Roger G.},
	month = jan,
	year = {2023},
	pages = {1},
}

@article{xi2025breaking,
  title={Breaking barriers in ICD classification with a robust graph neural network for hierarchical coding},
  author={Xi, Suyang and Shi, Jiesen and Yan, Jiachen and Lin, MingJing and Zhou, Xinyi and Cheng, Yuan and Ding, Hong and Kang, Chia Chao},
  journal={Scientific Reports},
  volume={15},
  number={1},
  pages={25676},
  year={2025},
  publisher={Nature Publishing Group UK London}
}

@article{li2020behrt,
  title={BEHRT: transformer for electronic health records},
  author={Li, Yikuan and Rao, Shishir and Solares, Jos{\'e} Roberto Ayala and Hassaine, Abdelaali and Ramakrishnan, Rema and Canoy, Dexter and Zhu, Yajie and Rahimi, Kazem and Salimi-Khorshidi, Gholamreza},
  journal={Scientific reports},
  volume={10},
  number={1},
  pages={7155},
  year={2020},
  publisher={Nature Publishing Group UK London}
}

@inproceedings{vaswani2017attention,
  title={Attention Is All You Need},
  author={Vaswani, Ashish and Shazeer, Noam and Parmar, Niki and Uszkoreit, Jakob and Jones, Llion and Gomez, Aidan N. and Kaiser, {\L}ukasz and Polosukhin, Illia},
  booktitle={Advances in Neural Information Processing Systems},
  year={2017}
}

@inproceedings{kipf2017semi,
  title={Semi-Supervised Classification with Graph Convolutional Networks},
  author={Kipf, Thomas N. and Welling, Max},
  booktitle={International Conference on Learning Representations},
  year={2017}
}

@article{rasmy2021medbert,
  title={Med-BERT: pretrained contextualized embeddings on large-scale structured electronic health records for disease prediction},
  author={Rasmy, Laila and Xiang, Yang and Xie, Ziqian and Tao, Cui and Zhi, Degui},
  journal={npj Digital Medicine},
  volume={4},
  number={1},
  pages={86},
  year={2021},
  publisher={Nature Publishing Group}
}

@inproceedings{choi2017gram,
  title={GRAM: Graph-based Attention Model for Healthcare Representation Learning},
  author={Choi, Edward and Bahadori, Mohammad Taha and Schuetz, Andy and Stewart, Walter F. and Sun, Jimeng},
  booktitle={Proceedings of the 23rd ACM SIGKDD International Conference on Knowledge Discovery and Data Mining},
  pages={787--795},
  year={2017}
}

@article{shang2019gbert,
  title={Pre-training of Graph Augmented Transformers for Medication Recommendation},
  author={Shang, Junyuan and Ma, Tengfei and Xiao, Cao and Sun, Jimeng},
  journal={arXiv preprint arXiv:1906.00346},
  year={2019}
}

@article{perotte2014diagnosis,
  title={Diagnosis code assignment: models and evaluation metrics},
  author={Perotte, Adler and Pivovarov, Rimma and Natarajan, Karthik and Weiskopf, Nicole and Wood, Frank and Elhadad, No{\'e}mie},
  journal={Journal of the American Medical Informatics Association},
  volume={21},
  number={2},
  pages={231--237},
  year={2014},
  publisher={BMJ Publishing Group}
}

@inproceedings{song2019medical,
  title={Medical Concept Embedding with Multiple Ontological Representations.},
  author={Song, Lihong and Cheong, Chin Wang and Yin, Kejing and Cheung, William K and Fung, Benjamin CM and Poon, Jonathan},
  booktitle={IJCAI},
  volume={19},
  pages={4613--4619},
  year={2019}
}

@article{ostrominski2021coding,
  title={Coding variation and adherence to methodological standards in cardiac research using the National Inpatient Sample},
  author={Ostrominski, John W and Amione-Guerra, Javier and Hernandez, Brian and Michalek, Joel E and Prasad, Anand},
  journal={Frontiers in cardiovascular medicine},
  volume={8},
  pages={713695},
  year={2021},
  publisher={Frontiers Media SA}
}

@inproceedings{choi2016doctor,
  title={Doctor ai: Predicting clinical events via recurrent neural networks},
  author={Choi, Edward and Bahadori, Mohammad Taha and Schuetz, Andy and Stewart, Walter F and Sun, Jimeng},
  booktitle={Machine learning for healthcare conference},
  pages={301--318},
  year={2016},
  organization={PMLR}
}

@article{choi2016retain,
  title={Retain: An interpretable predictive model for healthcare using reverse time attention mechanism},
  author={Choi, Edward and Bahadori, Mohammad Taha and Sun, Jimeng and Kulas, Joshua and Schuetz, Andy and Stewart, Walter},
  journal={Advances in neural information processing systems},
  volume={29},
  year={2016}
}

@inproceedings{ma2017dipole,
  title={Dipole: Diagnosis prediction in healthcare via attention-based bidirectional recurrent neural networks},
  author={Ma, Fenglong and Chitta, Radha and Zhou, Jing and You, Quanzeng and Sun, Tong and Gao, Jing},
  booktitle={Proceedings of the 23rd ACM SIGKDD international conference on knowledge discovery and data mining},
  pages={1903--1911},
  year={2017}
}

@inproceedings{choi2016med2vec,
  author    = {Edward Choi and Mohammad Taha Bahadori and Elizabeth Searles and Catherine Coffey and Michael Thompson and James Bost and Javier Tejedor-Sojo and Jimeng Sun},
  title     = {Multi-layer Representation Learning for Medical Concepts},
  booktitle = {Proceedings of the 22nd ACM SIGKDD International Conference on Knowledge Discovery and Data Mining},
  pages     = {1495--1504},
  year      = {2016},
  doi       = {10.1145/2939672.2939823}
}

@article{delong1988comparing,
  title={Comparing the Areas Under Two or More Correlated Receiver Operating Characteristic Curves: A Nonparametric Approach},
  author={DeLong, Elizabeth R. and DeLong, David M. and Clarke-Pearson, Daniel L.},
  journal={Biometrics},
  volume={44},
  number={3},
  pages={837--845},
  year={1988}
}

@article{sun2014fast,
  title={Fast implementation of DeLong’s algorithm for comparing the areas under correlated receiver operating characteristic curves},
  author={Sun, Xu and Xu, Weichao},
  journal={IEEE Signal Processing Letters},
  volume={21},
  number={11},
  pages={1389--1393},
  year={2014},
  publisher={IEEE}
}

@article{Pollard2018,
  author = {Pollard, Tom J. and Johnson, Alistair E. W. and Raffa, Jesse D. and Celi, Leo A. and Mark, Roger G. and Badawi, Omar},
  title = {The eICU Collaborative Research Database, a freely available multi-center database for critical care research},
  journal = {Scientific Data},
  year = {2018},
  volume = {5},
  pages = {180178},
  month = {Sep},
  doi = {10.1038/sdata.2018.178},
  pmid = {30204154},
  pmcid = {PMC6132188}
}

@article{goldberger2000physiobank,
  author  = {Goldberger, Ary L. and Amaral, Luis A. N. and Glass, Leon and
             Hausdorff, Jeffrey M. and Ivanov, Plamen Ch. and Mark, Roger G. and
             Mietus, Joseph E. and Moody, George B. and Peng, Chung-Kang and
             Stanley, H. Eugene},
  title   = {PhysioBank, PhysioToolkit, and PhysioNet: Components of a New Research Resource for Complex Physiologic Signals},
  journal = {Circulation},
  volume  = {101},
  number  = {23},
  pages   = {e215--e220},
  year    = {2000},
  month   = jun,
  doi     = {10.1161/01.CIR.101.23.e215}
}

@inproceedings{poulain2024graph,
  author    = {Poulain, Raphael and Beheshti, Rahmatollah},
  title     = {Graph Transformers on EHRs: Better Representation Improves Downstream Performance},
  booktitle = {The Twelfth International Conference on Learning Representations (ICLR)},
  year      = {2024},
  url       = {https://openreview.net/forum?id=pe0Vdv7rsL}
}

@article{choi2018mime,
  title={Mime: Multilevel medical embedding of electronic health records for predictive healthcare},
  author={Choi, Edward and Xiao, Cao and Stewart, Walter and Sun, Jimeng},
  journal={Advances in neural information processing systems},
  volume={31},
  year={2018}
}

@inproceedings{rupp2023exbehrt,
  title={Exbehrt: Extended transformer for electronic health records},
  author={Rupp, Maurice and Peter, Oriane and Pattipaka, Thirupathi},
  booktitle={International workshop on trustworthy machine learning for healthcare},
  pages={73--84},
  year={2023},
  organization={Springer}
}

@article{kraljevic2024foresight,
  title={Foresight—a generative pretrained transformer for modelling of patient timelines using electronic health records: a retrospective modelling study},
  author={Kraljevic, Zeljko and Bean, Dan and Shek, Anthony and Bendayan, Rebecca and Hemingway, Harry and Yeung, Joshua Au and Deng, Alexander and Balston, Alfred and Ross, Jack and Idowu, Esther and others},
  journal={The Lancet Digital Health},
  volume={6},
  number={4},
  pages={e281--e290},
  year={2024},
  publisher={Elsevier}
}

@article{renc2024zero,
  title={Zero shot health trajectory prediction using transformer},
  author={Renc, Pawel and Jia, Yugang and Samir, Anthony E and Was, Jaroslaw and Li, Quanzheng and Bates, David W and Sitek, Arkadiusz},
  journal={NPJ digital medicine},
  volume={7},
  number={1},
  pages={256},
  year={2024},
  publisher={Nature Publishing Group UK London}
}

@inproceedings{pang2021cehr,
  title={CEHR-BERT: Incorporating temporal information from structured EHR data to improve prediction tasks},
  author={Pang, Chao and Jiang, Xinzhuo and Kalluri, Krishna S and Spotnitz, Matthew and Chen, RuiJun and Perotte, Adler and Natarajan, Karthik},
  booktitle={Machine learning for health},
  pages={239--260},
  year={2021},
  organization={PMLR}
}

@inproceedings{ma2018kame,
  title={Kame: Knowledge-based attention model for diagnosis prediction in healthcare},
  author={Ma, Fenglong and You, Quanzeng and Xiao, Houping and Chitta, Radha and Zhou, Jing and Gao, Jing},
  booktitle={Proceedings of the 27th ACM international conference on information and knowledge management},
  pages={743--752},
  year={2018}
}

@book{who_icd10,
  author    = {{World Health Organization}},
  title     = {International Statistical Classification of Diseases and Related Health Problems, 10th Revision},
  publisher = {World Health Organization},
  year      = {2016},
  edition   = {5th},
  address   = {Geneva}
}

@inproceedings{berg2012empirical,
  title={An empirical investigation of statistical significance in NLP},
  author={Berg-Kirkpatrick, Taylor and Burkett, David and Klein, Dan},
  booktitle={Proceedings of the 2012 joint conference on empirical methods in natural language processing and computational natural language learning},
  pages={995--1005},
  year={2012}
}

\newpage

\appendix

\section{HICD-Graph Hierarchy:  Example}
\label{app:hicd_graph_example}

\begin{figure}[t]
    \centering
    \includegraphics[width=0.8\linewidth]{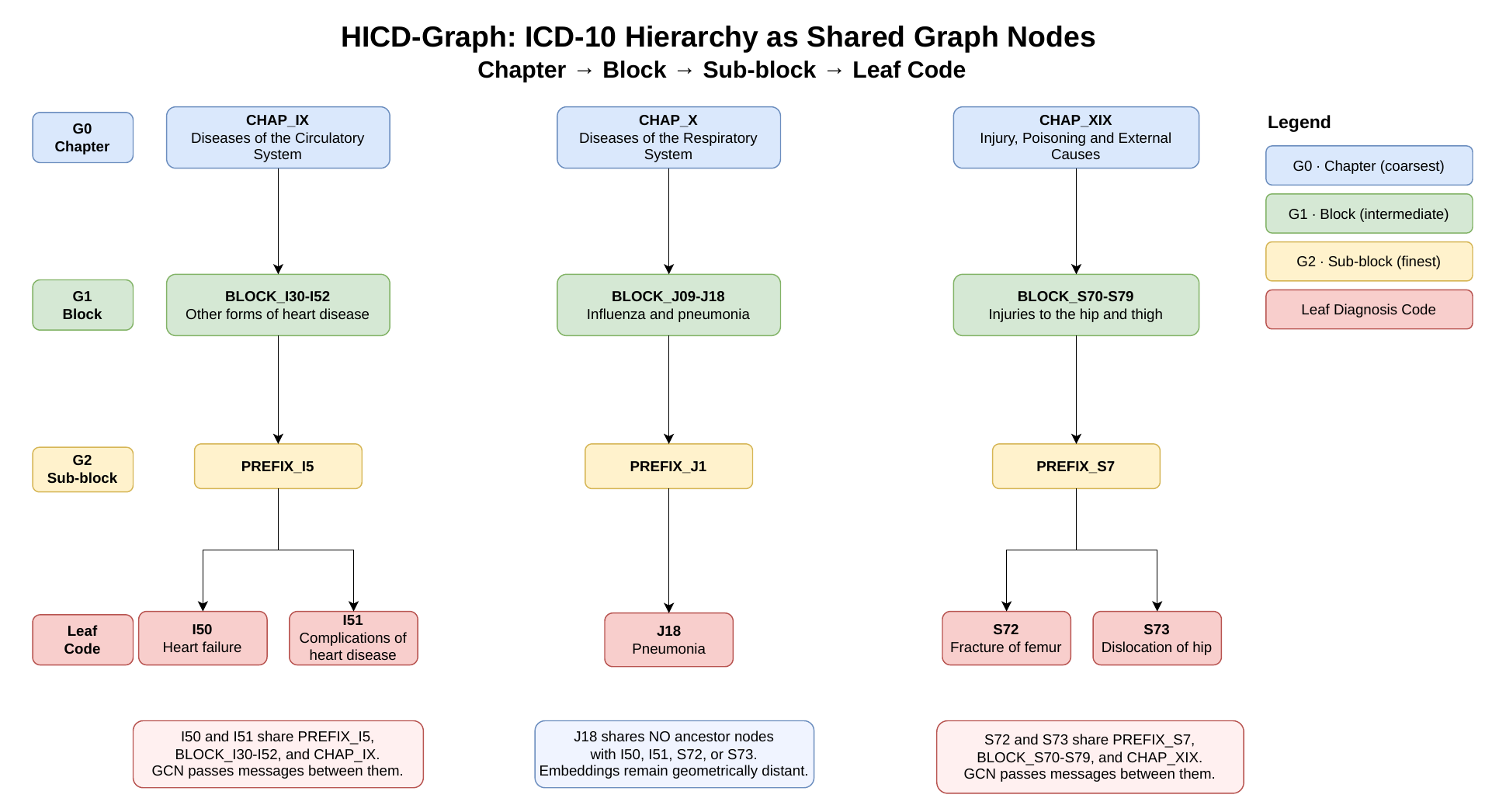}
    \caption{Example of the \hicdgraph\ hierarchy subgraph.}
    \label{fig:hicd_graph_example}
\end{figure}

Figure~\ref{fig:hicd_graph_example} illustrates how two leaf diagnosis codes share hierarchy nodes in \hicdgraph. Consider \texttt{I50.21} (\emph{Acute systolic (congestive) heart failure}) and \texttt{I50.32} (\emph{Chronic diastolic (congestive) heart failure}). Both leaves connect, via weighted edges, to the same \texttt{G2} node (\texttt{I5}$x$ prefix group), the same \texttt{G1} node (official block \texttt{I30--I52}, \emph{Other forms of heart disease}), and the same \texttt{G0} node (Chapter~IX, \emph{Diseases of the circulatory system}). Shared ancestors propagate information between clinically related leaves during message passing, while unrelated leaves remain separated in the hierarchy subgraph.

\begin{figure}[h]
    \centering
    \includegraphics[width=\linewidth]{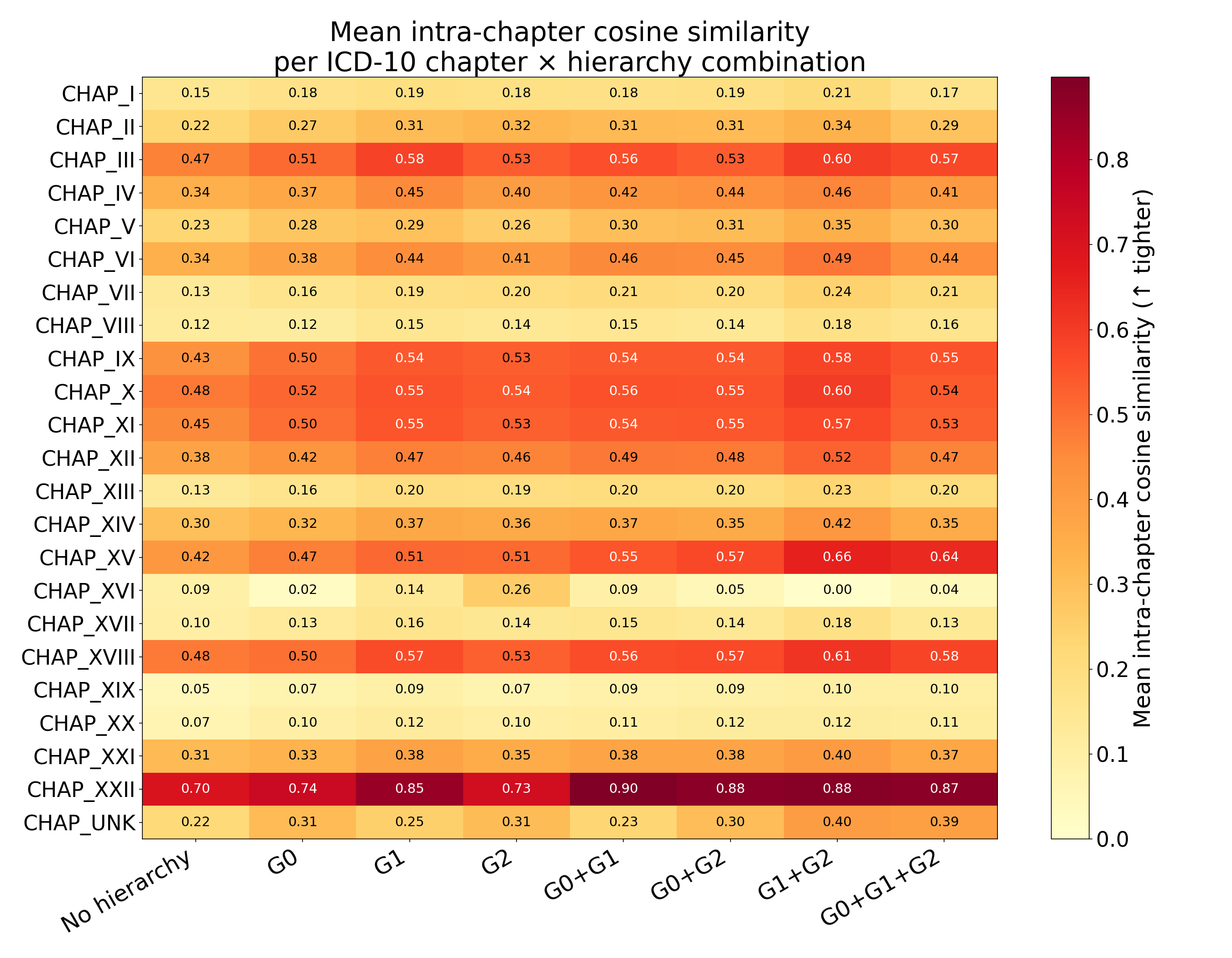}
    \caption{Mean intra-chapter cosine similarity among ICD-10 leaf codes in our pretraining data of MIMIC-IV, computed across all eight hierarchy configurations (\texttt{G0}/\texttt{G1}/\texttt{G2} ablations). Each row corresponds to one ICD-10 chapter, and each column to a hierarchy configuration ordered as in the main results tables. Higher values (darker) indicate that codes within the same chapter are embedded more tightly together. Hierarchy increases similarity across most chapters, with the largest gains in diverse chapters such as Chapter~XIX (\emph{Injury and poisoning}, 5{,}937 codes).}
    \label{fig:intra_chapter_cosine}
\end{figure}

\section{Implementation Details}
\label{app:implementation}

\subsection{\hicdbert\ Architecture and Training}
\label{app:hicdbert_details}

\hicdbert\ extends a BEHRT-style encoder by summing diagnosis-token, age,
position, and segment embeddings, with optional additive embeddings for the
three hierarchy levels \texttt{G0}, \texttt{G1}, and \texttt{G2}. The contextualized sequence is
processed by a stack of Transformer blocks, and the pretraining objective is
masked language modeling (MLM). For downstream prediction, the encoder is
initialized from the pretrained checkpoint and a task-specific multilayer
perceptron is applied to the final \texttt{[CLS]} representation.

\subsubsection{Architecture}
\hicdbert\ builds on a BEHRT-style Transformer encoder and augments the standard
token, age, position, and segment embeddings with up to three hierarchy-aware
embedding channels corresponding to \texttt{G0}, \texttt{G1}, and \texttt{G2}. These hierarchy
embeddings are added to the input representation and are learned jointly with
the rest of the model. For downstream prediction, the contextualized
\texttt{[CLS]} representation is passed through a task-specific multilayer
perceptron classifier.

\begin{table}[h]
\centering
\small
\caption{\hicdbert\ architecture details.}
\label{tab:hicdbert_arch}
\begin{tabular}{ll}
\toprule
Component & Setting \\
\midrule
Backbone & BEHRT-style Transformer encoder \\
Embedding components & Token + age + position + segment + optional $G0/G1/G2$ embeddings \\
Hidden size ($d_{\text{model}}$) & 512 \\
Transformer layers & 6 \\
Attention heads & 8 \\
Feed-forward dimension & 2048 \\
Dropout & 0.1 \\
Maximum sequence length & 256 \\
MLM head & Linear $\rightarrow$ GELU $\rightarrow$ LayerNorm $\rightarrow$  Decoder \\
Task head & 3-layer MLP on \texttt{[CLS]} representation \\
\bottomrule
\end{tabular}
\end{table}

\subsubsection{MLM Pretraining}
We pretrain \hicdbert\ with a masked language modeling objective over diagnosis
sequences. The model is optimized to recover masked diagnosis tokens from their
clinical context while jointly conditioning on age, position, segment, and the
enabled hierarchy-level embeddings. Table~\ref{tab:hicdbert_pretrain}
summarizes the pretraining configuration.

\begin{table}[h]
\centering
\small
\caption{\hicdbert\ MLM pretraining hyperparameters.}
\label{tab:hicdbert_pretrain}
\begin{tabular}{ll}
\toprule
Hyperparameter & Value \\
\midrule
Train batch size & 256 \\
Validation batch size & 128 \\
Number of workers & 8 \\
Epochs & 100 \\
Optimizer & AdamW \\
Learning rate & $1\times10^{-4}$ \\
Weight decay & 0.01 \\
Scheduler & CosineAnnealingLR \\
Scheduler $T_{\max}$ & 100 \\
Minimum learning rate & $1\times10^{-6}$ \\
Loss & CrossEntropyLoss \\
\bottomrule
\end{tabular}
\end{table}

\subsubsection{Task Fine-Tuning}
For downstream tasks, we initialize the encoder from the best MLM checkpoint
and optimize a task-specific classification objective. Binary tasks use \texttt{BCEWithLogitsLoss} with a positive class weight
$w^{+} = N_{\text{neg}}^{\text{train}} / N_{\text{pos}}^{\text{train}}$, computed from the training split, to address label imbalance.
The best checkpoint is selected by
validation AUROC. For binary tasks, a decision threshold is swept over
$[0.01, 0.99]$ (200 steps) on the validation set and stored in the checkpoint
for test-set evaluation. Table~\ref{tab:hicdbert_finetune} lists the full
fine-tuning hyperparameters.

\begin{table}[h]
\centering
\small
\caption{\hicdbert\ downstream fine-tuning hyperparameters.}
\label{tab:hicdbert_finetune}
\begin{tabular}{ll}
\toprule
Hyperparameter & Value \\
\midrule
Train batch size & 128 \\
Validation batch size & 256 \\
Number of workers & 8 \\
Epochs & 64 \\
Optimizer & AdamW \\
Learning rate & $1\times10^{-5}$ \\
Weight decay & 0.01 \\
Scheduler & ReduceLROnPlateau (mode: max) \\
Scheduler patience & 2 \\
Scheduler factor & 0.5 \\
Early stopping patience & 8 \\
Early stopping min delta & $10^{-4}$ \\
Early stopping monitor & Validation AUROC \\
Model selection criterion & Best validation AUROC \\
Binary-task loss & BCEWithLogitsLoss ($w^{+} = N_{\text{neg}}/N_{\text{pos}}$) \\
Threshold selection & Sweep $[0.01, 0.99]$, 200 steps, max F1 on val set \\
\bottomrule
\end{tabular}
\end{table}

\subsection{\hicdgraph: Graph-Augmented Hierarchy Model}
\label{app:graph_details}

The graph-based model has two stages. First, we construct a diagnosis
co-occurrence graph over ICD roots using PMI-weighted edges, then augment the
graph with hierarchy edges corresponding to the enabled \texttt{G0}, \texttt{G1}, and \texttt{G2}
levels. Second, we train a two-layer GCN for node embedding learning and use
the resulting embeddings to initialize a patient-level Transformer.

\subsubsection{Graph Construction}
We construct an undirected diagnosis graph whose nodes correspond to ICD root
codes. Horizontal edges are weighted by PMI computed from within-admission
co-occurrence, and vertical edges are added to encode enabled hierarchy levels.
Table~\ref{tab:graph_construction} summarizes the graph construction settings.
\begin{table}[h]
\centering
\small
\caption{Diagnosis graph construction hyperparameters.}
\label{tab:graph_construction}
\begin{tabular}{ll}
\toprule
Hyperparameter & Value \\
\midrule
Diagnosis vocabulary & ICD-10-CM codes  \\
ICD-9 handling & Mapped to ICD-10-CM \\
Horizontal edges & PMI-weighted co-occurrence edges \\
PMI filtering & Positive PMI only \\
PMI cutoff & 52nd percentile \\
Hierarchy levels & \texttt{G0}, \texttt{G1}, \texttt{G2} ablated independently \\
Hierarchy edge weight & 1.0 \\
Graph type & Undirected weighted graph \\
\bottomrule
\end{tabular}
\end{table}

\subsubsection{GCN Embedding Training}
We learn node embeddings with a two-layer GCN trained using a link-prediction
objective over observed and sampled non-edge pairs. The resulting embeddings
are used to initialize the downstream patient encoder. The training
configuration is given in Table~\ref{tab:gcn_hparams}.
\begin{table}[h]
\centering
\small
\caption{GCN embedding training hyperparameters.}
\label{tab:gcn_hparams}
\begin{tabular}{ll}
\toprule
Hyperparameter & Value \\
\midrule
Input node feature dimension & 64 \\
Input feature initialization & Random normal \\
GCN layers & 2 \\
Hidden dimension & 128 \\
Output embedding dimension & 128 \\
Edge weights & Min-max normalized to $[0,1]$ \\
Training objective & Link prediction \\
Positive pairs & Existing graph edges \\
Negative pairs & Random non-edge pairs \\
Loss & BCEWithLogitsLoss \\
Optimizer & Adam \\
Learning rate & 0.005 \\
Epochs & 400 \\
\bottomrule
\end{tabular}
\end{table}

\subsubsection{Transformer Fine-Tuning on Graph Embeddings}
The learned graph embeddings initialize the diagnosis embedding table of a
patient-level Transformer operating over temporally ordered visits. Visit
representations are formed by masked mean pooling over code embeddings, and
patient representations are obtained through attention pooling across visits.
Table~\ref{tab:graph_transformer_hparams} provides the fine-tuning details.
\begin{table}[h]
\centering
\small
\caption{Patient-level Transformer hyperparameters for the graph-based model.}
\label{tab:graph_transformer_hparams}
\begin{tabular}{ll}
\toprule
Hyperparameter & Value \\
\midrule
Embedding initialization & GCN node embeddings \\
Embedding dimension & 128 \\
Maximum visits per patient & 10 \\
Code aggregation within visit & Masked mean pooling \\
Visit positional encoding & Learnable \\
Transformer layers & 2 \\
Attention heads & 4 \\
Feed-forward dimension & 256 \\
Dropout & 0.3 \\
Pooling across visits & Attention pooling \\
Classifier & Linear layer on pooled patient representation \\
Batch size & 64 \\
Epochs & 15 \\
Optimizer & AdamW \\
Learning rate & $2\times10^{-5}$ \\
Weight decay & $1\times10^{-5}$ \\
Scheduler & ReduceLROnPlateau (mode: max) \\
Scheduler patience & 2 \\
Scheduler factor & 0.5 \\
Model selection criterion & Best validation AUROC \\
Loss & BCEWithLogitsLoss ($w^{+} = N_{\text{neg}}/N_{\text{pos}}$) \\
Class balancing & WeightedRandomSampler + $w^{+}$ \\
Threshold selection & Sweep $[0.01, 0.99]$, 200 steps, max F1 on val set \\
Gradient clipping & Max norm 1.0 \\
\bottomrule
\end{tabular}
\end{table}

\subsection{Statistical Testing}
\label{app:statistical_testing}

For all pairwise comparisons of AUROC between a hierarchy-augmented variant and
the no-hierarchy baseline evaluated on the same test set, we use the paired
DeLong test~\citep{delong1988comparing}, which provides a closed-form test for
the difference between two correlated AUROC estimates. Our implementation was
verified against the reference implementation of Sun and
Xu~\citep{sun2014fast}\footnote{\url{https://github.com/yandexdataschool/roc\_comparison}},
with all p-values agreeing to machine precision ($<10^{-15}$) across all
comparisons. We report significance at the $p<0.05$ threshold.

\subsection{Hierarchy Ablations}
\label{app:ablation_settings}

For both \hicdbert\ and the graph-augmented model, we evaluate all eight
hierarchy settings:
\[
(0,0,0),\ (0,0,1),\ (0,1,0),\ (0,1,1),\
(1,0,0),\ (1,0,1),\ (1,1,0),\ (1,1,1).
\]
Here, $(0,0,0)$ corresponds to the non-hierarchical baseline, while
$(1,1,1)$ uses the full hierarchy. In \hicdbert\, these settings determine
which hierarchy embeddings are added to the token representation. In the
graph-based model, they determine which hierarchy edges are added to the
diagnosis graph and which ancestor nodes are included during sequence
tokenization.

\end{document}